\documentclass[letterpaper]{article} 
\usepackage{aaai2026}  
\usepackage{times}  
\usepackage{helvet}  
\usepackage{courier}  
\usepackage[hyphens]{url}  
\usepackage{graphicx} 
\usepackage{natbib}  
\usepackage{caption} 
\usepackage{graphicx}
\usepackage{cite}
\usepackage{color}
\usepackage{amssymb}
\usepackage{booktabs}
\usepackage{pifont}
\usepackage{amsmath}
\usepackage{multirow}
\usepackage{makecell}
\usepackage{algorithm}
\usepackage{algorithmic}

\usepackage{newfloat}
\usepackage{listings}
\DeclareCaptionStyle{ruled}{labelfont=normalfont,labelsep=colon,strut=off} 
\floatstyle{ruled}
\newfloat{listing}{tb}{lst}{}
\floatname{listing}{Listing}
\title{RoadSceneVQA: Benchmarking Visual Question Answering in Roadside Perception Systems for Intelligent Transportation System}
\author {
    Runwei Guan\textsuperscript{\rm 1,2,$*$},
    Rongsheng Hu\textsuperscript{\rm 3},
    Shangshu Chen\textsuperscript{\rm 4},
    Ningyuan Xiao\textsuperscript{\rm 1},
    Xue Xia\textsuperscript{\rm 1}, 
    Jiayang Liu\textsuperscript{\rm 1},
    Beibei Chen\textsuperscript{\rm 5},
    Ziren Tang\textsuperscript{\rm 1},
    Ningwei Ouyang\textsuperscript{\rm 6},
    Shaofeng Liang\textsuperscript{\rm 1},
    Yuxuan Fan\textsuperscript{\rm 1}, \\
    Wanjie Sun\textsuperscript{\rm 7},
    Yutao Yue\textsuperscript{\rm 1,2,}\thanks{\textit{CA}: \{runwayrwguan, yutaoyue\}@hkust-gz.edu.cn}
}

\affiliations {
    \textsuperscript{\rm 1}Information Hub, The Hong Kong University of Science and Technology (Guangzhou), Guangzhou, China \\
    \textsuperscript{\rm 2}Institute of Deep Perception Technology, JITRI, Wuxi, China\\ 
    \textsuperscript{\rm 3}School of Artificial Intelligence and Computer Science, Jiangnan University, Wuxi, China \\
     \textsuperscript{\rm 4}National Research Center of Cultural Industries, Central China Normal University, Wuhan, China \\
     \textsuperscript{\rm 5}School of Marxism, Jilin University, Changchun, China \\  
     \textsuperscript{\rm 6}School of Advanced Technology, Xi'an Jiaotong-Liverpool University, Suzhou, China \\
     \textsuperscript{\rm 7}School of Remote Sensing and Information Engineering, Wuhan University, Wuhan, China \\
}

\usepackage{bibentry}

\begin{document}
\maketitle


\begin{abstract}
Current roadside perception systems mainly focus on instance-level perception, which fall short in enabling interaction via natural language and reasoning about traffic behaviors in context. To bridge this gap, we introduce RoadSceneVQA, a large-scale and richly annotated visual question answering (VQA) dataset specifically tailored for roadside scenarios. The dataset comprises 34,736 diverse QA pairs collected under varying weather, illumination, and traffic conditions, targeting not only object attributes but also the intent, legality, and interaction patterns of traffic participants. RoadSceneVQA challenges models to perform both explicit recognition and implicit commonsense reasoning, grounded in real-world traffic rules and contextual dependencies. To fully exploit the reasoning potential of Multi-modal Large Language Models (MLLMs), we further propose CogniAnchor Fusion (CAF), a vision-language fusion module inspired by human-like scene anchoring mechanisms. Moreover, we propose the Assisted Decoupled Chain-of-Thought (AD-CoT) to enhance the reasoned thinking via CoT prompting and multi-task learning. Based on the above, we propose the baseline model RoadMind. Experiments on RoadSceneVQA and CODA-LM benchmark show that the pipeline consistently improves both reasoning accuracy and computational efficiency, allowing the MLLM to achieve state-of-the-art performance in structural traffic perception and reasoning tasks.
\end{abstract}

\begin{links}
    \link{Datasets}{https://github.com/GuanRunwei/RS-VQA}
\end{links}


\section{Introduction}
\label{sec:intro}

\begin{figure*}
\centering
\includegraphics[width=0.998\linewidth]{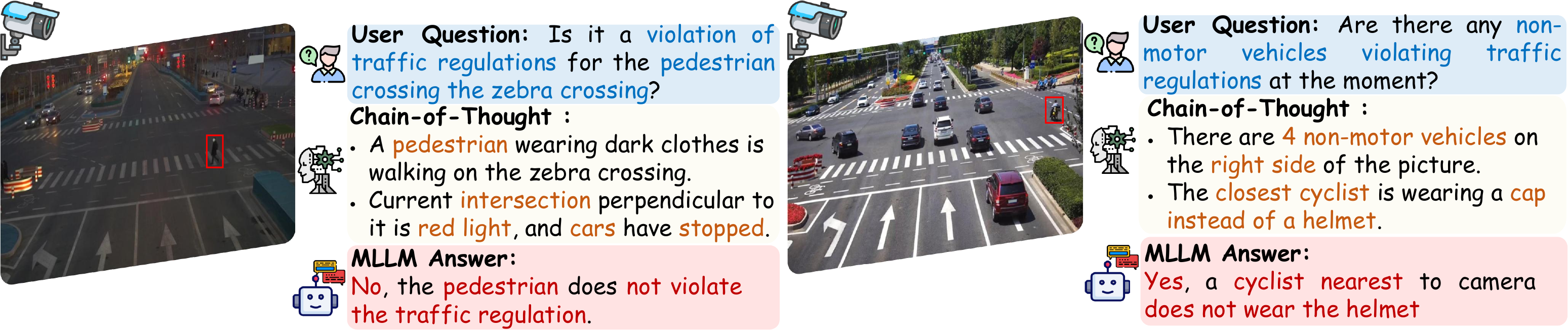}
\caption{Overview of RoadSceneVQA with samples.}
\label{fig:overview}
\end{figure*}

With the advancement of Multi-modal Large Language Models (MLLMs), the capabilities for perception and reasoning in traffic scenarios have significantly improved \cite{zhou2024vision}. As a critical component of traffic perception, roadside perception offers a distinct advantage over vehicle-based perception by providing a top-down perspective that enables clearer observation of the states and behaviors of traffic participants, and a more comprehensive understanding of the overall scene \cite{bejarbaneh2024exploring}.

However, current roadside perception systems primarily focus on automated tasks such as detection \cite{guan2023man}, tracking \cite{liang2025cognitive}, trajectory prediction \cite{dai2025large}, and traffic flow forecasting \cite{huang2023v2x}. While these approaches have achieved a certain degree of success, they largely lack human-in-the-loop perception, which mostly emphasizes instance-level recognition, but falls short in event-level or holistic scene-level understanding. Moreover, these systems often exhibit limited scalability and interpretability in complex environments, and lack the flexibility to recognize unforeseen objects and events. Recently, Vision-Language Models (VLMs), empowered by large-scale pretraining, have demonstrated strong reasoning capabilities \cite{guan2025referring} and generalization across a wide range of multi-modal tasks \cite{sun20243d,sun2025city,zhang2025open3dvqa}. These models have shown promising performance in vehicle-based perception. Moreover, the reasoning ability of MLLMs, enabled by prompting, allows for rapid adaptation to novel traffic events.

\begin{table*}
    \setlength\tabcolsep{3.3pt}
    \centering
    \begin{tabular}{cc|ccccc}
    \hline
      \textbf{Datasets} & \textbf{Venues} & \textbf{QA Generation} & \textbf{QA/Caption items} & \textbf{Scenes} & \textbf{Reasoning} & \textbf{Domains} \\
    \hline
      Talk2Car \cite{deruyttere2019talk2car} & EMNLP$_{2019}$ & Manual & 2.4k & 2.4k & \ding{55} & Driving \\
      NuScenes-QA \cite{qian2024nuscenes}  & AAAI$_{2024}$ & Template & 83.3k & 0.9k & \ding{55} & Driving\\
      DriveLM \cite{sima2024drivelm} & ECCV$_{2024}$ & Template + Manual & 15.4k & 188k & \ding{51} & Driving \\
      Talk2Radar \cite{guan2024talk2radar} & ICRA$_{2025}$ & Manual & 8.7k & 8.7k & \ding{55} & Driving\\
      nuPrompt \cite{wu2025language} & AAAI$_{2025}$ & LLM & 40.8k & 0.9k & \ding{55} & Driving \\
      DriveBench \cite{xie2025vlms} & ICCV$_{2025}$ & Template + Manual & 20.5k & - & \ding{51} & Driving \\
      TUM-VideoQA \cite{zhou2025tumtraffic} & ICML$_{2025}$ & LLM + Manual & 87.3k & 1k & \ding{55} & Roadside\\
    \hline
      \textbf{RoadSceneVQA (ours)} & \textbf{2025} & \textbf{LLM + Manual} & \textbf{34.7k} & \textbf{26} & \ding{51} & \textbf{Roadside} \\
    \hline     
    \end{tabular}
    \caption{Summary of vision-language-based datasets oriented at traffic scenarios.}
    \label{tab:dataset_compare}
\end{table*}

While VLM-based roadside perception benchmarks have achieved remarkable progress in visual grounding or image captioning \cite{zhou2025tumtraffic,guan2024findvehicle}, their reliance on explicit localization fundamentally limits higher-level traffic understanding. Existing benchmarks measure perceptual accuracy, yet fail to evaluate whether models comprehend implicit traffic regulations, such as determining ``Is any pedestrian violating traffic rules?", which requires synthesizing signal states, spatial contexts, and behavioral dynamics. This semantic gap persists because localization tasks disregard causal relationships (e.g., a vehicle may be in a crosswalk but not violating if signals permit). 

Building upon the aforementioned, we construct RoadSceneVQA, a large-scale Visual Question Answering (VQA) dataset tailored for roadside perception. Leveraging the rich object distributions and contextual diversity of the impressive Rope3D dataset \cite{ye2022rope3d}, which provides roadside detection data across various traffic scenarios. It covers a wide range of traffic events, object states, lighting conditions, and weather environments. RoadSceneVQA includes both explicit and implicit question-answer pairs centered on traffic scene reasoning, posing challenges to models in terms of scene understanding and commonsense reasoning. More importantly, we introduce an agile Collaborative Human-Machine Annotation (CH-MA) system to enable high-quality and scalable annotation, which is critical for constructing this comprehensive VQA dataset.

Last but not least, current MLLMs typically adopt a simple token-level concatenation strategy for the fusion of visual and language tokens. However, this design introduces two limitations: \textbf{(1)} Irrelevant visual tokens, such as background noise, are forcibly mixed with key textual tokens, disrupting the model’s ability to accurately localize objects;  \textbf{(2)} Imbalanced information interaction, where dominant visual features overwhelm the textual signals, impairs the model’s ability to follow language instructions \cite{yang2024mma}. To enhance MLLMs’ contextual awareness for commonsense reasoning, we draw inspiration from human vision-language collaborative cognition and propose a plug-and-play visual-language fusion module, termed CogniAnchor Fusion (CAF). CAF pre-anchors potential regions of interest and enables the LLM to reason precisely and efficiently. Moreover, advanced MLLM such as GPT-4o \cite{hurst2024gpt} have demonstrated strong capabilities in knowledge-driven reasoning, and have been applied to autonomous driving for Chain-of-Thought (CoT) generation, enabling more stable and interpretable reasoning paradigms \cite{liao2025cot}. To address the challenges of complex roadside scenarios, we propose a novel method called Assisted Decoupled Chain-of-Thought (AD-CoT), designed to support reliable scene understanding and reasoning on lightweight edged MLLMs (smaller than 8B). Building on these, we further introduce our baseline MLLM, RoadMind, specifically tailored for multi-modal roadside reasoning.

In summary, our contributions are as follows,

\begin{enumerate}
    \item \textbf{RoadSceneVQA}, a large-scale VQA dataset designed for roadside perception, which comprises 34,736 samples of Q-A pair, spanning perception and reasoning tasks.
    \item \textbf{CH-MA}, an agile human–machine collaborative annotation framework for VQA datasets, which improves annotation efficiency while enhancing accuracy and quality.
    \item \textbf{CogniAnchor Fusion}, a visual-language fusion module inspired by human scene cognition mechanisms, tailored for MLLMs. It enhances the reasoning capabilities while maintaining high computational efficiency.
    \item \textbf{RoadMind}, a multi-modal large language model en-powered by \textbf{Assisted Decoupled Chain-of-Thought (AD-CoT)} for roadside traffic perception and reasoning.
\end{enumerate}

\begin{figure*}
\centering
\includegraphics[width=0.98\linewidth]{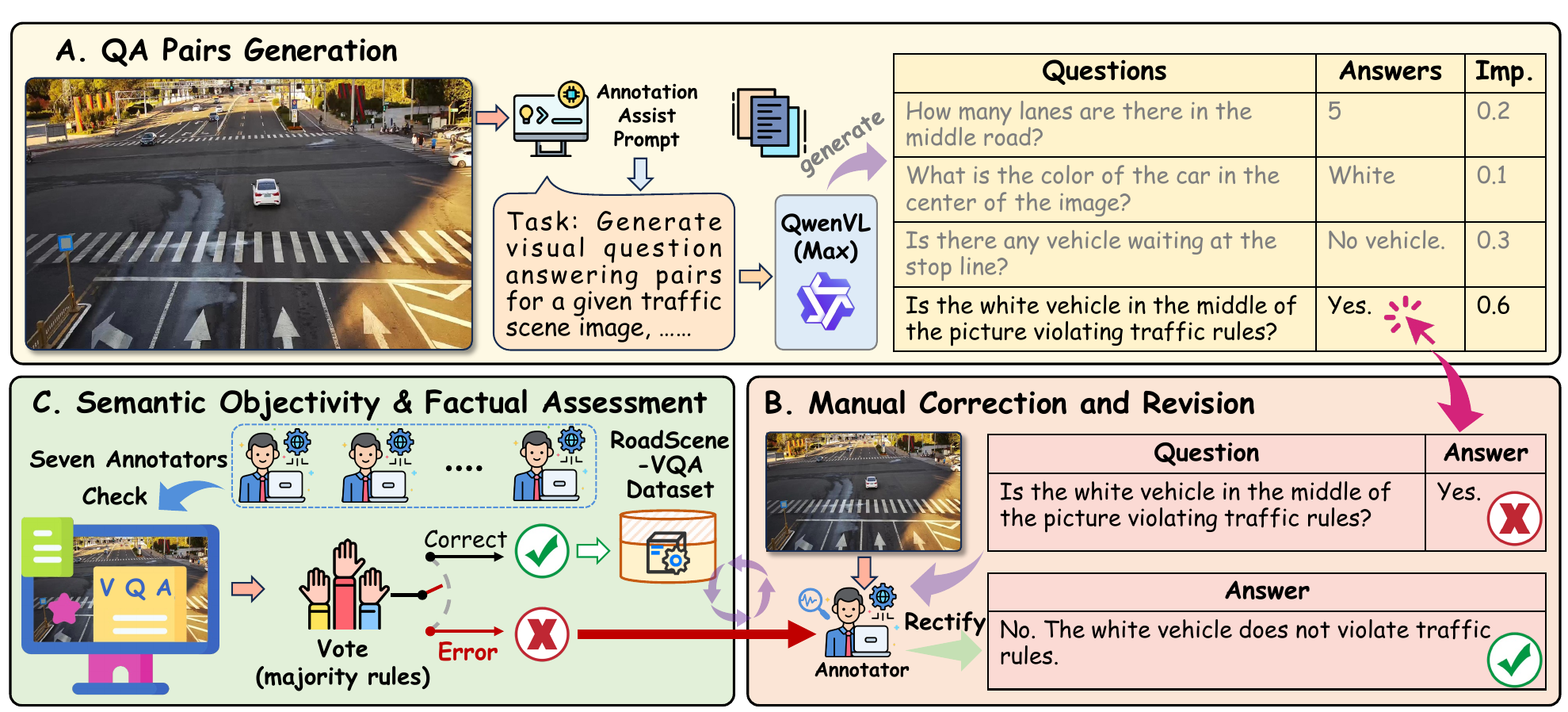}
\caption{Annotation of RoadSceneVQA dataset by the agile Collaborative Human-Machine Annotation system.}
\label{fig:anno_process}
\end{figure*}

\section{Related Works}
\label{sec:related}

\textbf{Open-Ended VQA based on MLLMs.} Visual Question Answering (VQA) has significantly evolved from conventional selection-based and multi-class frameworks to open-ended, human-like systems \cite{kim2025visual}. Early VQA models treat the task as multi-class classification, predicting answers from predefined sets by fusing CNN-extracted visual features with textual embeddings \cite{yang2016stacked,antol2015vqa,malinowski2015ask}. Despite promising results, their fixed answer sets hinder handling complex queries and open-ended reasoning.
With the rise of large-scale datasets like Visual Genome \cite{krishna2017visual}, researchers adopt encoder-decoder architectures. Incorporating attention mechanisms, the models enable more free-form text generation \cite{nam2017dual,anderson2018bottom}. However, semantic coherence and factual accuracy remain challenges due to difficulties in grounding visual concepts with language.
In the advent of MLLMs, by integrating pre-trained knowledge with visual understanding, MLLMs generate natural, open-ended responses similar to human answers \cite{sima2024drivelm,zhang2025mpdrive}.
Currently, VQA in traffic scenarios primarily focuses on explicit QA \cite{qian2024nuscenes,wu2025language}, with limited efforts exploring context-aware reasoning, which are constrained to on-vehicle settings \cite{xie2025vlms}. Existing roadside VQA benchmarks tend to emphasize object localization, typically outputting bounding boxes \cite{zhou2025tumtraffic}, which is difficult to reason effectively about the behavior of traffic participants or to draw decision-level conclusions. 
To address these limitations, we propose RoadSceneVQA, which fills this gap by implicitly incorporating commonsense knowledge into VQA tasks for traffic, enabling more holistic and actionable scene understanding.

\noindent \textbf{Roadside Perception for ITS.} Roadside perception is vital for Intelligent Transportation Systems (ITS), offering a complementary perspective to on-board vehicle sensors for traffic monitoring and decision-making. Recent advances have addressed a range of tasks. Object detection \cite{ye2022rope3d,yu2022dair} and semantic segmentation \cite{muhammad2022vision} from roadside cameras have been explored to identify traffic participants in varying conditions. Occupancy-based methods aim to construct bird’s-eye-view representations for scene understanding \cite{chang2023bev}. Moreover, traffic flow prediction models enable proactive traffic management and congestion control \cite{xu2021continuous}. Despite these advances, most existing researches focus on perception only or fixed outputs such as bounding boxes, which often require post-processing or human intervention for scene understanding. In contrast, VQA offers a natural and flexible interface for reasoning about traffic scenes, yet current VQA datasets are predominantly vehicle-centric and lack the complexity of roadside contexts. To bridge this gap, we propose RoadSceneVQA, a comprehensive VQA dataset tailored for roadside perception. It introduces diverse, context-rich, and cognitively challenging question-answer pairs covering traffic understanding, behavior reasoning, and commonsense inference, paving the way for next-generation multi-modal reasoning in ITS applications.

\section{RoadSceneVQA Dataset}
\label{sec:dataset}
RoadSceneVQA presents the first vision-language benchmark that shifts traffic intelligence evaluation from perceptual recognition to regulation-aware cognitive reasoning from roadside view. Unlike existing VQA datasets which predominantly query explicit object attributes or rely on video temporal cues (e.g., `Is the vehicle speeding?'), RoadSceneVQA pioneers image-based compliance judgment, demanding models to answer questions like `Is the pedestrian violating traffic rules given current signal states and crosswalk topology?'. This necessitates a tripartite integration of: (i) visual-semantic grounding (e.g., correlating pedestrian position with signal phase), (ii) regulatory knowledge internalization, and (iii) counterfactual causality (e.g., inferring `Would the cyclist have violated if the light were green?'). In this section, we first introduce \textbf{agile Collaborative Human–Machine Annotation system}, a user-friendly and efficient framework for VQA annotation. Then, we provide a statistical analysis of the dataset across multiple dimensions.

\begin{figure*}
\centering
\includegraphics[width=0.99\linewidth]{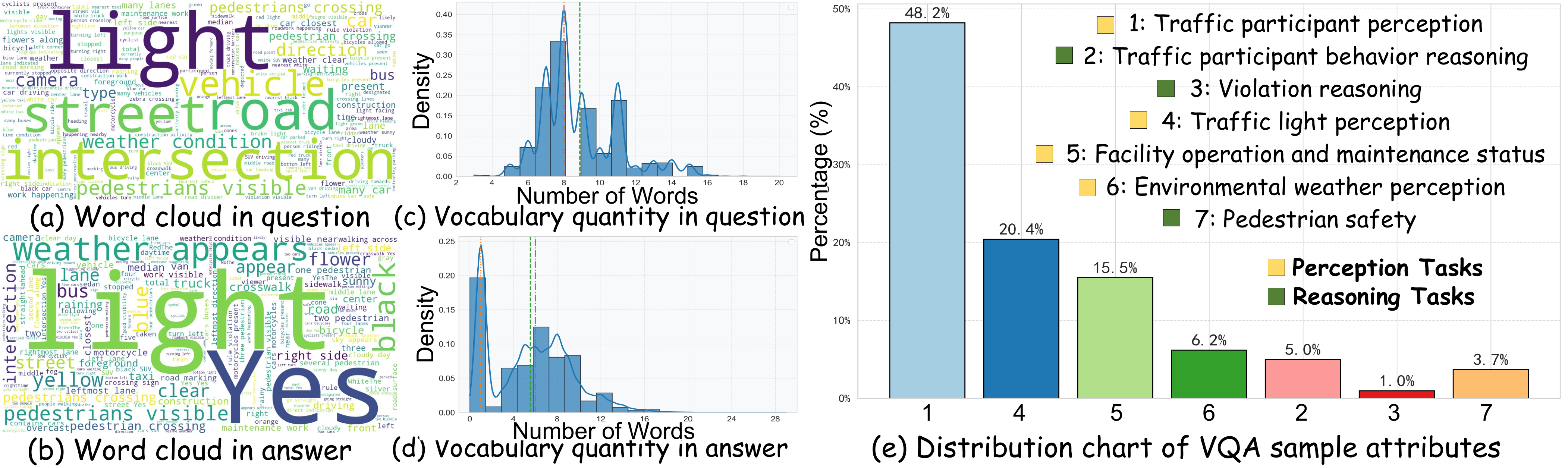}
\caption{Statistics of RoadSceneVQA dataset.}
\label{fig:statis}
\end{figure*}

\subsection{Agile Collaborative Human-Machine Annotation}
\label{subsec:agile}
To construct a natural and accurate question-answering dataset that emulates human-like reasoning, we propose an agile Collaborative Human–Machine Annotation system (CH-MA). This framework is designed to address several key challenges that often compromise annotation quality: \textbf{(1)} Manual question writing is prone to subjective bias; \textbf{(2)} Isolated focus on individual traffic participants, without considering broader traffic context, fails to reflect realistic scenarios; \textbf{(3)} Question diversity heavily depends on the expertise of annotators, which may lead to inconsistencies; \textbf{(4)} Template-based QA generation often lacks naturalness and suffers from limited semantic expressiveness in real-world interaction settings. To tackle these issues, our CH-MA framework is structured into three stages, each designed to balance automation and human expertise for scalable, high-quality VQA dataset construction.

\textbf{Stage A.} To assist annotators in expanding reasoning space and exploring contextual diversity, we introduce QwenVL-Max \cite{bai2025qwen2}, a state-of-the-art MLLM. The annotation leader provides QwenVL-Max with a tailored prompt (see supple. materials) and the current roadside traffic image. The model then generates four candidate QA pairs, each accompanied by an estimated reasoning contribution score for the traffic scene. Annotators review the image and the generated QA pairs, assess their factual alignment (subjectively determined) and reasoning contribution, and select the highest-quality pair as the initial reference.

\textbf{Stage B.} Given the inherent limitations and hallucination tendencies of MLLMs, the QA pairs generated in \textbf{Stage A} are not guaranteed to be factually accurate or contextually precise. In this stage, annotators are required to revise and refine the selected QA pair to ensure factual correctness, contextual alignment, and linguistic clarity, while maintaining a formal and neutral tone.

\textbf{Stage C.} To minimize annotation errors caused by subjectivity or carelessness, a final quality control stage is conducted. All annotated samples are reviewed by a panel of seven annotators, and only those receiving majority approval are admitted into the dataset. Samples that do not pass this validation are sent back to Stage B for further refinement, ensuring high annotation quality and consistency.

\subsection{Dataset Statistics}
\label{subsec:statistics}

As shown in Table \ref{tab:dataset_compare}, RoadSceneVQA dataset contains a total of 34.7K QA pairs. In contrast to TUMTraffic-VideoQA \cite{zhou2025tumtraffic}, RoadSceneVQA emphasizes reasoning-based questions that incorporate real-world commonsense and domain-specific knowledge. Figure \ref{fig:statis} presents a quantitative overview of RoadSceneVQA, including a word cloud of the QA corpus, vocabulary length distributions for questions and answers, and the attribute distribution of VQA samples. It is observed that the question and answer texts in RoadSceneVQA exhibit a broad distribution with rich vocabulary coverage. The dataset captures a wide variety of perception, understanding, and reasoning questions tailored for traffic scenarios, while preserving a domain gap that reflects real-world complexity and enhances generalization.

\begin{figure*}
\centering
\includegraphics[width=0.998\linewidth]{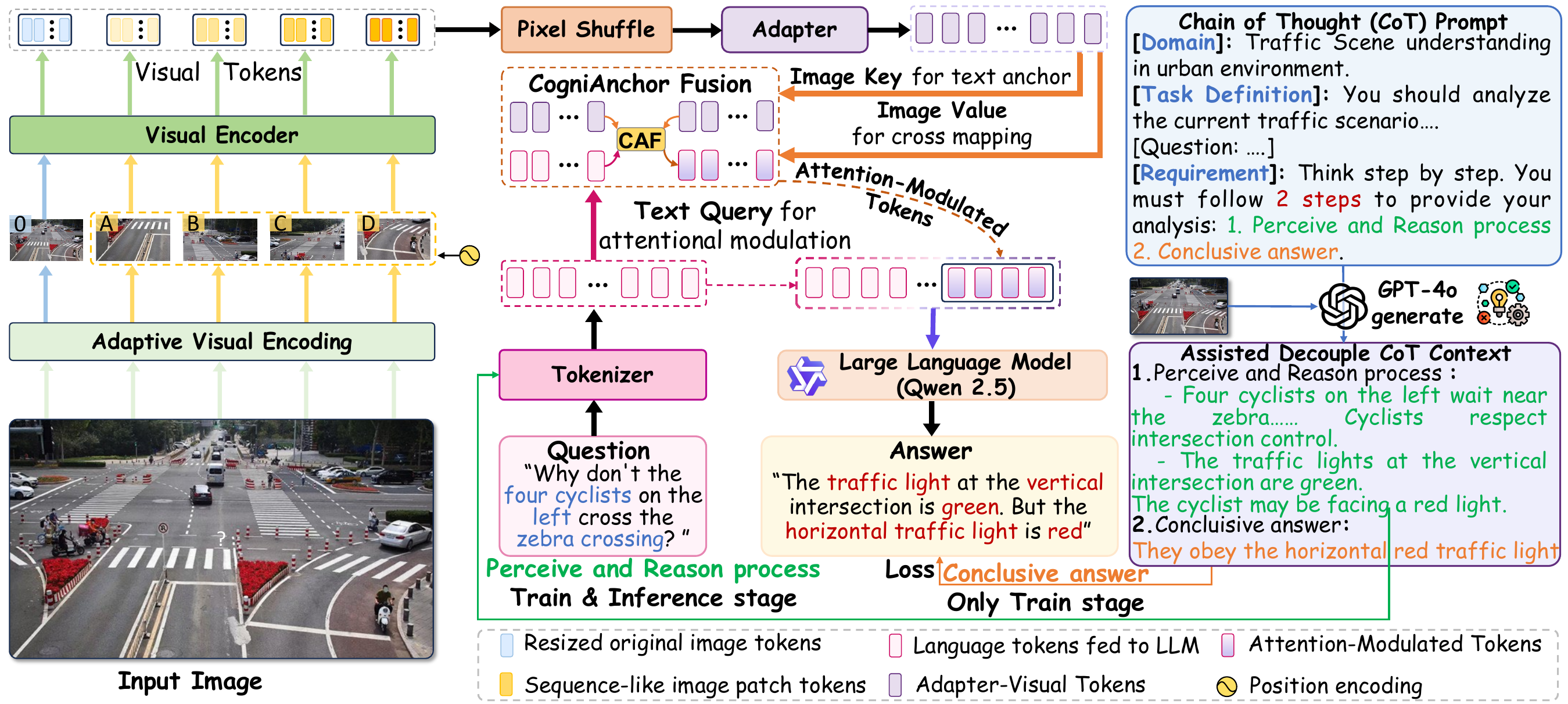}
\caption{The architecture of RoadMind Model, where we propose CogniAnchor Fusion and Assisted Decoupled CoT.}
\label{fig:model}
\end{figure*}

\section{Methodology}

We propose RoadMind, a MLLM tailored for roadside traffic perception and reasoning, incorporating two key components: \textbf{(1) CAF}, a cognitively inspired, language-driven pre-anchoring strategy that guides visual attention toward critical regions based on human perception mechanisms; (2) \textbf{AD-CoT}, a reasoning prompting framework to support reliable and interpretable inference on edge-deployed models.

\subsection{Overall Pipeline}
As illustrated in Figure \ref{fig:model}, given an input image $\mathbf{I} \in \mathbb{R}^{H \times W \times 3}$ captured from a roadside camera, we first apply adaptive visual encoding module, which decomposes the image into a sequence of visual patches while simultaneously preserving a downsampled version of the original image to retain global context. In parallel, a globally pooled representation $\mathbf{I_g} \in \mathbb{R}^{H' \times W' \times 3}$ is extracted by downsampling $\mathbf{I}$, which serves as a global prior for subsequent reasoning. Formally, the encoding process produces:

\begin{equation}
    \mathbf{S} = \{\mathbf{p_1}, \mathbf{p_2}, \ldots, \mathbf{p_N}, \mathbf{I_g}\}, \quad \mathbf{p_i} \in \mathbb{R}^{d},
\end{equation}
where $\mathbf{S}$ is the set of patch embeddings, and each $\mathbf{p_i}$ corresponds to a visual token derived from local regions of $\mathbf{I}$. Subsequently, the patch sequence $\mathbf{S}$ is processed by a Visual Encoder (InternViT) \cite{chen2024internvl}, which extracts high-level representations. To further enhance feature density, we apply the Pixel Shuffle operation and obtain the final visual feature $\mathbf{V}$. For the question $\mathbf{T}$, we leverage GPT-4o \cite{hurst2024gpt} to generate a Chain-of-Thought prompt $\mathbf{T}^{OC}$ and a structured answer $\mathbf{T}^{SA}$, both conditioned on the image $\mathbf{I}$ and the original question $\mathbf{T}$. We then concatenate the $\mathbf{T}$ with its generated $\mathbf{T}^{CoT}$ and obtain $\mathbf{T}^{OC}$:

\begin{equation}
    \mathbf{T}^{OC} = \mathtt{Concat}(\mathbf{T}, \mathbf{T}^{CoT}).
\end{equation}

This enriched textual input $\mathbf{T}^{OC}$ is tokenized and encoded into text embeddings. Meanwhile, the visual features extracted from InternViT are passed through a lightweight MLP adapter to align their representation space with that of the text. The aligned image features and the encoded textual representation are then fused via the CogniAnchor Fusion module, resulting in text-conditioned visual tokens. These tokens are concatenated with the language embeddings and jointly fed into Qwen 2.5, to produce the final answer $\mathbf{A}$:

\begin{equation}
    \mathbf{A} = \mathtt{Qwen2.5}(\mathtt{Concat}(\mathtt{CAF}(\mathbf{V}, \mathbf{T}^{OC}), \mathbf{T}^{OC}))
\end{equation}

\subsection{CongiAnchor Fusion}
Most MLLMs adopt a naive token-level concatenation that merges visual and textual tokens. However, this paradigm requires the model to learn cross-modal spatial correspondences from scratch, often causing inefficient grounding of visual evidence performance. To address these limitations, we propose CogniAnchor Fusion (CAF), a cognitively inspired fusion mechanism that mirrors the typical human process of answering visual questions, which obtains language-anchored candidate visual regions for the following fusion with language features.
Besides, to ensure efficiency, CAF is formulated as a linear-based cross attention. However, linear attention often suffers from non-injectivity, which leads to semantic ambiguity and impair the model's ability to accurately understand candidate regions within complex scenes. Therefore, CAF is inspired by InLine Attention \cite{han2024bridging}, enabling scalable integration with existing MLLMs while maintaining efficiency in reasoning.

Given the textual input as $\mathbf{Q}^{T} \in \mathbb{R}^{N \times d}$ for the source driving the query. $\{\mathbf{K}^{I}, \mathbf{V}^I \} \in \mathbb{R}^{N \times d}$ are the key and value upon visual features to calculate the attention matrix and form the representation. For the query vector $\mathbf{Q_i}^T$, the function of attention weight $\mathtt{AW}(\cdot)$ is defined as:

\begin{equation}
    \begin{aligned}
       & \mathtt{AW}(\mathbf{Q_i}^T, \mathbf{K}) = [\mathtt{\phi}(\mathbf{Q_i}^T)^{\top}\phi(\mathbf{K_1}), \dots, \mathtt{\phi}(\mathbf{Q_i}^T)^{\top}\phi(\mathbf{K_N})]^{\top} \\
    & \quad \quad \quad \quad \quad \quad - \frac{1}{N}\sum^N_{s=1} \phi(\mathbf{Q_i}^T)^{\top} \phi(\mathbf{K_s}) + \frac{1}{N},
    \end{aligned}
\end{equation}
where $\phi(\cdot)$ is the kernel function. $\frac{1}{N}\sum^N_{s=1} \phi(\mathbf{Q_i}^T)^{\top} \phi(\mathbf{K_s})$ indicates the averaging operation over all similarity scores. $\frac{1}{N}$ is a constant term added to ensure that attention weights sum to 1 (meet the normalization). Finally, the language-driven image attention features can be formulated as:

\begin{equation}
    \begin{aligned}
        & \mathtt{AW}(\mathbf{Q_i}^T, \mathbf{K}^I)^\top \mathbf{V}^I = \phi(\mathbf{Q_i}^T)^\top [\sum^N_{j=1}\phi(\mathbf{K_j}^I)V_j^\top] \\
        & - [\phi(\mathbf{Q_i}^T)^\top \sum_{j=1}^N \phi(\mathbf{K_j}^I) - 1] \frac{1}{N} \sum_{j=1}^{N} \mathbf{V_j}^I.
    \end{aligned}
\end{equation}

Here, by adjusting the calculation order (aggregating key-value pairs first and then interacting with the query), the complexity is reduced from $O(N^2)$ of traditional Softmax attention to $O(N)$.

In conclusion, CAF enhances the causality of the generated responses, leveraging the reasoning capability of a large MLLM to empower a smaller and deployable MLLM.

\begin{table*}
    \setlength\tabcolsep{1.4pt}
    \centering
    \begin{tabular}{c|c|c|ccccccc}
    \hline
       \textbf{Models} & \textbf{LLM} & \textbf{Params} & \textbf{Exact Match} & \textbf{ROUGH-L} & \textbf{BLEU-4} & \textbf{METEOR} & \textbf{CIDEr} & \textbf{SPICE}& \textbf{GPT-Score} \\
    \hline
    MiniCPM-o 2.6 (nsft) & LLaMA3 & 8B & 0.021 & 0.193 & 0.080 & 0.324 & 0.661 & 0.124 & 0.428 \\
    \hline
    EM-VLM4AD & T5-Large & 0.8B & 0.086 & 0.249 & 0.134 & 0.299 & 1.142 & 0.148 & 0.372 \\
    InternVL3 & Qwen 2.5 & 0.9B & 0.142 & \textbf{0.386} & 0.166 & \textbf{0.385} & 1.656 & 0.170 & 0.403 \\
    TinyLLaVA & OpenELM  & 0.9B & 0.098 & 0.283 & 0.127 & 0.319 & 1.150 & 0.147 & 0.394 \\
    \textbf{RoadMind} & Qwen 2.5 & 0.9B & \textbf{0.144} & 0.372 & \textbf{0.179} & 0.322 & \textbf{1.867} & \textbf{0.188} & \textbf{0.440} \\
    \hline
    MobileVLM v2 & MobileLLaMA & 1.7B & 0.081 & 0.296 & 0.137 & 0.338 & 1.399& 0.162 & 0.417 \\
    InternVL3 & Qwen 2.5 & 2B & \textbf{0.151} & 0.403 & 0.155 & 0.315 & \textbf{1.834} & 0.201 & 0.465 \\
    MiniCPM-V 2.0 & MiniCPM & 2.8B & 0.136 & 0.389 & 0.149 & \textbf{0.352} & 1.717 & 0.194 & 0.477 \\
    \textbf{RoadMind} & Qwen 2.5 & 2B & 0.142 & \textbf{0.405} & \textbf{0.158} & 0.347 & 1.705 & \textbf{0.219} & \textbf{0.489} \\
    \hline
    Qwen2.5-VL & Qwen2.5 & 7B & 0.152 & 0.403 & 0.159 & 0.385 & 1.689 & 0.213 & 0.497\\
    InternVL3 & Vicuna & 8B & \textbf{0.161} & 0.411 & 0.162 & 0.398 & 1.735 & 0.208 & 0.532 \\
    MiniCPM-o 2.6 & LLaMA3 & 8B & 0.147 & 0.388 & \textbf{0.165} & 0.394 & 1.826 & 0.199 & 0.527 \\
    \textbf{RoadMind} & Qwen 2.5 & 8B & 0.157 & \textbf{0.425} & 0.157 & \textbf{0.411} & \textbf{1.836} & \textbf{0.221} & \textbf{0.554} \\

    \hline
    \end{tabular}
    \caption{Overall performances on RoadSceneVQA dataset. nsft: No Supervised Fine-Tuning.}
    \label{tab:benchmark_compare}
\end{table*}

\subsection{Assisted Decoupled Chain-of-Thought (AD-CoT)}
Due to the limited capacity of lightweight roadside MLLMs, their reasoning capabilities are inherently constrained. To address this, we propose AD-CoT that combines knowledge distillation with cognitive transfer, where GPT-4o serves as a soft supervisory prior to enhance the model’s ability to perform context-aware reasoning in complex traffic scenarios. Exactly, given an input image and a question, we first feed the image along with a CoT prompt into GPT-4o. The model generates an assisted reasoning context, which includes both a Perceive and Reason Process and a Conclusive Answer, where the former is then concatenated with the original question and passed into our lightweight model, RoadMind, as enriched input. Meanwhile, the conclusive answer from GPT-4o is paired with the human-annotated ground truth answer to construct a multi-task learning objective, enabling RoadMind to simultaneously learn both reasoning logic and answer accuracy.
The loss function $L_{\text{MTL}}$ is presented as follows:
\begin{equation}
    \begin{aligned}
        & L_{\text{MTL}} = \frac{1}{\sigma_{\text{hard}}^2}[\sum^L_{l=1}\mathtt{log}p(\mathbf{y}_l^{\text{hard}}|\mathbf{y}_{<l}^{\text{hard}}, \mathbf{x}, \mathbf{q})] + \\
        &  \frac{1}{\sigma_{\text{soft}}^2}[\sum^{\text{min}(L, L^{'})}_{l=1}\mathtt{D_{\text{KL}}}(p_l^{\mathtt{GPT}}||\hat{p}_l)] + \mathtt{log}\sigma_{\mathtt{hard}} + \mathtt{log}{\sigma_{\text{soft}}},
    \end{aligned}
\end{equation}
where $\mathbf{x}$ denotes the image input while $\mathbf{q}$ is question context. $\mathbf{y}^{\text{hard}}$ is the ground truth answer. $L$ is the length of ground truth answer while $L^{'}$ is the length of GPT-4o's answer. $\mathtt{D_{\text{KL}}}$ is Kullback-Leibler divergence function. $p^{\text{GPT}}_l$ is the probability distribution of GPT-4o at position $l$ while $\hat{p}_l$ is the probability distribution of the model at position $l$. $\sigma_{\text{hard}}$ and $\sigma_{\text{soft}}$ are two learnable weights for uncertainty modelling. 

\begin{figure}
\centering
\includegraphics[width=0.98\linewidth]{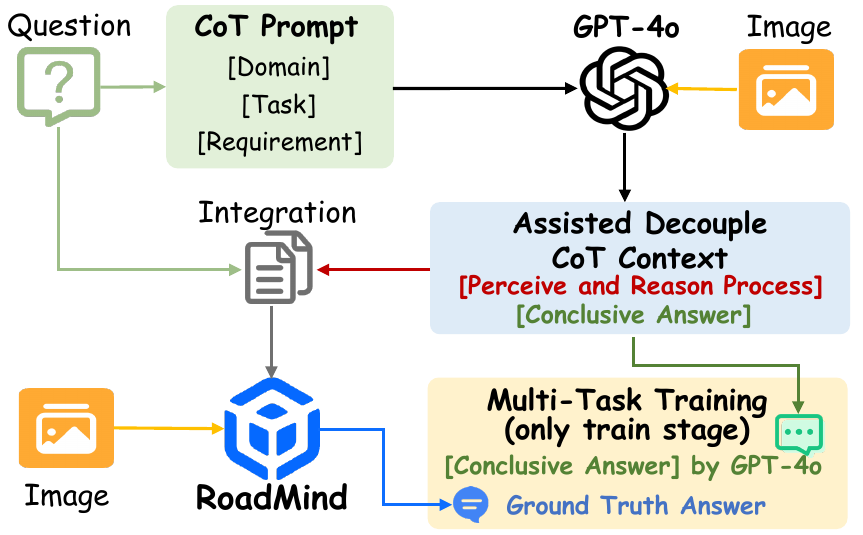}
\caption{The pipeline of Assisted Decouple CoT.}
\label{fig:cot}
\end{figure}

\begin{figure*}
\centering
\includegraphics[width=0.998\linewidth]{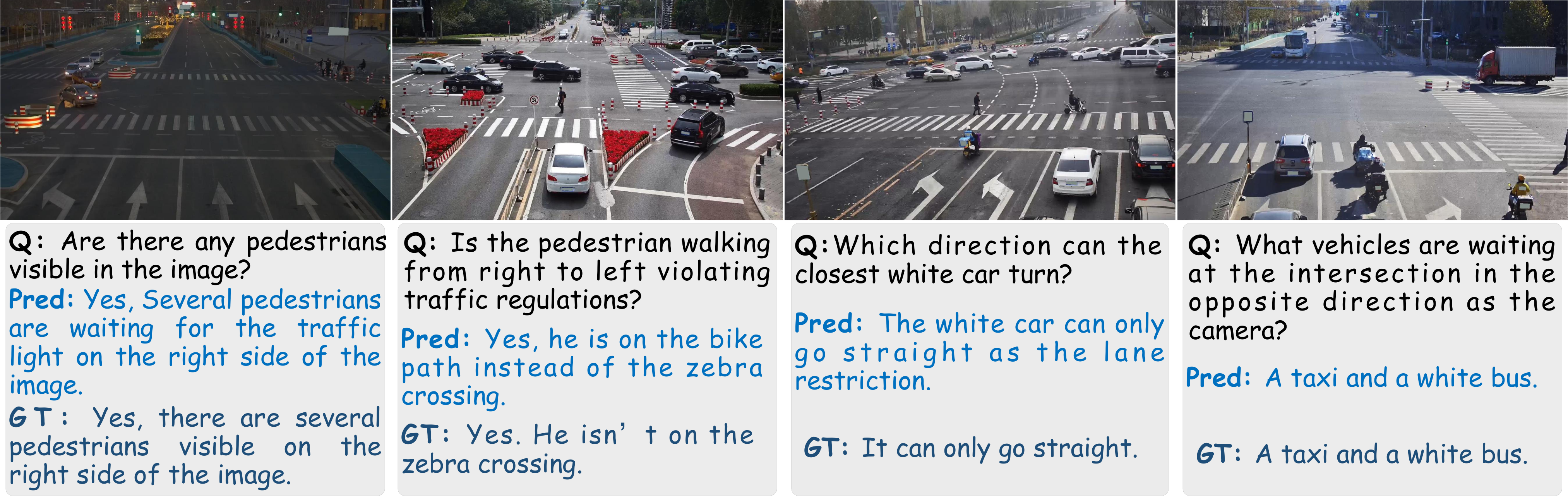}
\caption{Inference results upon RoadMind-8B.}
\label{fig:vis}
\end{figure*}

\section{Experiments}
\subsection{Settings of Experiments}
\noindent \textbf{Dataset.} Our proposed RoadSceneVQA is adopted for baseline experiments, including 30,058 samples for training and 4,677 samples for test. To evaluate the generalization of models, CODA-LM \cite{chen2025automated}, another VQA dataset for ego-car VQA is also included, containing 20,495 samples in train set while 2,123 samples in test set.

\noindent \textbf{Models.} Besides RoadMind models (0.9B, 2.0 and 8.0B), QwenVL \cite{bai2025qwen2}, InternVL \cite{chen2024internvl}, MiniCPM series \cite{yao2024minicpm}, MobileVLM v2 \cite{chu2024mobilevlm}, TinyLLaVA \cite{zhou2024tinyllava} and EM-VLM4AD \cite{gopalkrishnanmulti} are included, containing containing MLLM for general and specific fields. Besides, for the proposed CogniAnchor Fusion, we set the head number of attention as 8.

\noindent \textbf{Training and Evaluation.} For the training of RoadSceneVQA, we resize the image as 448 $\times$ 448 (px). We finetune the pretrained models for one epoch with the initial learning rate of 1e-5. We freeze the visual encoder while unfreeze LLM and MLP projector. We set the max sequence length as 16384. We adopt AdamW optimizer and set the weight decay as 0.05 with the cosine scheduler and warm-up ratio of 0.03. We train models on 4 A100 GPUs with the batch size per GPU as 1. For CODA-LM, we keep the settings consistent with \cite{chen2025automated}. We include Exact Match, ROUGH-L, \cite{lin2004rouge}, BLEU-4 \cite{papineni2002bleu}, METEOR \cite{meteor}, CIDEr \cite{vedantam2015cider}, SPICE \cite{anderson2016spice} and GPT-Score \cite{sima2024drivelm} as evaluation metrics, which are widely used in text generation field.

\subsection{Quantitative Results}
\noindent \textbf{Overall.} Table \ref{tab:benchmark_compare} presents the performance of multi-scale MLLMs on the RoadSceneVQA. Overall, across three model size tiers, proposed RoadMind consistently achieves the best performance on most evaluation metrics. Notably, it excels in the GPT-Score, indicating stronger capabilities in contextual understanding and reasoning, as well as generating responses that are more aligned with human preferences. Furthermore, RoadMind (0.9B) outperforms both the zero-shot MiniCPM-o 2.6 (8B) and the fully fine-tuned MobileVLM v2 (1.7B) in terms of GPT-Score. This shows that our design, particularly the CAF module and the AD-CoT strategy, effectively enhancing the reasoning ability of lightweight models. These enable them to reach or surpass the performance of significantly larger counterparts.

\begin{table}
\centering
\setlength\tabcolsep{3.2pt}
\small
\begin{tabular}{c|cccc}
\hline
    \textbf{Sample Types} & \textbf{ROUGH-L}  & \textbf{METEOR} & \textbf{CIDEr} & \textbf{GPT-Score} \\
\hline
    \multicolumn{5}{c}{\textbf{Perception}} \\
\hline
    \textbf{TPP} & 0.404 & 0.391 & 1.894 & 0.561\\
    \textbf{TLP} & 0.417 & 0.348 & 1.877 & 0.583\\
    \textbf{EWP} & 0.486 & 0.437 & 1.929 & 0.601\\
    \textbf{FOMS} & 0.432 & 0.408 & 1.752 & 0.530\\
\hline
    \multicolumn{5}{c}{\textbf{Reasoning}} \\
\hline
    \textbf{VR} & 0.218 & 0.177 & 1.376 & 0.338\\
    \textbf{TPBR} & 0.323 & 0.216 & 1.438 & 0.429\\
    \textbf{PS} & 0.287 & 0.253 & 1.393 & 0.371\\
\hline
\end{tabular}
\caption{Performances upon RoadMind-8B. The abbreviations in \textbf{Sample Types} are derived from \textbf{Figure \ref{fig:statis}(e)}.}
\label{tab:sample_type_experiment}
\end{table}

\noindent \textbf{Sample Types.} Table \ref{tab:sample_type_experiment} reports the performance of RoadMind-8B across different sample categories. The model performs significantly better on perception-based questions compared to reasoning-based ones, highlighting the relative difficulty of high-level contextual inference. Within the perception category, the model achieves higher accuracy on weather and traffic light recognition, whereas its performance on understanding traffic participants and roadside infrastructure remains comparatively lower.

\noindent \textbf{Generalization.} Table \ref{tab:coda_lm_experiments} shows that RoadMind-8B achieves strong performance on the CODA-LM benchmark, even surpassing InternVL 1.5-20B, which highlights the remarkable capability of RoadMind across diverse reasoning tasks.

\begin{table}
\setlength\tabcolsep{3.0pt}
\small
    \centering
    \begin{tabular}{c|c|cccc|c}
    \hline
    \multirow{2}{*}{\textbf{Models}} & \multirow{2}{*}{\textbf{GTS$\uparrow$}} & \multicolumn{4}{c}{\underline{\textbf{Regional Perception  $\uparrow$}}} & \multirow{2}{*}{\textbf{STS$\uparrow$}} \\
     &  & \textbf{All} & \textbf{Vehicle} & \textbf{VRU} & \textbf{Sign}  \\
    \hline
    MiniGPT-v2-7B & 11.58 & 15.93 & 18.74 & 13.58 & 15.71 & 10.0 \\
    Shikra-7B & 12.24 & 22.94 & 28.29 & 17.88 & 20.0 & 10.2 \\
    QwenVL2-7B & 18.22 & 26.62 & 35.48 & 24.16 & 20.86 & 22.06 \\
    MiniCPM2.5-8B & 41.12 & 57.2 & 61.91 & 54.82 & \textbf{59.43} & 48.48 \\
    InternVL1.5-20B & 38.38 & 61.53 & 63.77 & 53.14 & 50.57 & 41.18 \\
    \hline
    \textbf{RoadMind-8B} & \textbf{48.50} & \textbf{70.65} & \textbf{74.25} & \textbf{59.78} & 47.43 & \textbf{54.28} \\
    \hline
    \end{tabular}
    \caption{Comparison of models on CODA-LM dataset. \textbf{GTS:} General Text-Score; \textbf{STS:} Suggestion Text-Score.}
    \label{tab:coda_lm_experiments}
\end{table}

\begin{table}
\centering
\setlength\tabcolsep{1.8pt}
\begin{tabular}{c|c|c|c|c}
\hline
    \textbf{NSFT} &  \textbf{LoRA}  & \textbf{SFT} & \textbf{SFT + CAF} & \textbf{SFT + CAF + AD-CoT} \\
\hline
    0.428 & 0.452 & 0.527 & 0.533 & \textbf{0.549} \\
\hline
\end{tabular}
\caption{Migration performance (GPT-Score) on MiniCPM-o 2.6 for CAF and AD-CoT.}
\label{tab:transfer_performance}
\end{table}

\begin{table}
\centering
\setlength\tabcolsep{1.0pt}
\small
\begin{tabular}{c|cc|ccc}
\hline
    \textbf{VL Fusion} & \textbf{Params} & \textbf{FLOPs}  & \textbf{ROUGH-L} & \textbf{METEOR} & \textbf{SPICE} \\
\hline
    \textbf{CAF+Concat} & 0.924K & 61.08M  & \textbf{0.425} & 0.411 & \textbf{0.221}\\
\hline
    SCA+Concat & \textbf{0.768K} & \textbf{18.79M} & 0.389 & 0.401 & 0.197\\
    LCA+Concat & 1.049M & 371.86M & 0.397 & 0.386 & 0.201\\
    CA+Concat & 1.063M & 495.41M & 0.418 & \textbf{0.422} & 0.217 \\
    Concat & - & -  & 0.366 & 0.397 & 0.187\\
    Image as query & - & - & 0.404 & 0.374 & 0.193 \\
\hline
\end{tabular}
\caption{Ablation experiments on approaches of vision-language token fusion in RoadMind-8B.(SCA: \cite{guan2025watervg}, LCA: \cite{wang2020linformer}, CA: \cite{vaswani2017attention})}
\label{tab:vl_fusion_ablation}
\end{table}

\subsection{Ablation Studies}
\textbf{First}, Table \ref{tab:transfer_performance} presents migration performance of MiniCPM-o 2.6 for CAF and AD-CoT, indicating the impressive transfer ability of these two approaches. \textbf{Second}, Table \ref{tab:vl_fusion_ablation} presents the superior performances of CAF, with the trade-off between precision and computational cost. \textbf{Third}, Table \ref{tab:cot_ablation} shows each settings of AD-CoT matter while AD-CoT performs better than MCoT \cite{zhang2024multimodal}.

\begin{table}[H]
\centering
\setlength\tabcolsep{3.0pt}
\small
\begin{tabular}{c|cc|cc}
\hline
    \multirow{2}{*}{\textbf{CoT}} & \multicolumn{2}{c}{\textbf{Perception}} &  \multicolumn{2}{c}{\textbf{Reasoning}} \\
    \cmidrule(lr){2-3} \cmidrule(lr){4-5}
    & METEOR & GPT-Score & METEOR & GPT-Score \\
    \cmidrule(lr){1-5} 
    \textbf{AD-CoT} & 0.420 & \textbf{0.568} & \textbf{0.339} & \textbf{0.445} \\
    \hline
    Only OQ Input & 0.392 & 0.523 & 0.301 & 0.401 \\
    Only GT (train) & \textbf{0.428} & 0.549 & 0.325 & 0.439 \\
    Only CA (train) & 0.387 & 0.491 & 0.331 & 0.395 \\
    \hline
    MCoT & 0.422 & 0.536 & 0.319 & 0.431 \\
\hline
\end{tabular}
\caption{Ablation studies in CoT. OQ: original question; GT: ground truth; CA: conclusive answer (from GPT-4o).}
\label{tab:cot_ablation}
\end{table}

\subsection{Visualization Results}
Figure \ref{fig:vis} presents qualitative reasoning results from RoadMind-8B. The results show that RoadMind is capable of accurately perceiving contextual scene elements and target states, and generating rule-compliant, context-aware responses in accordance with traffic regulations.

\section{Conclusion}
We propose RoadSceneVQA, a large-scale VQA dataset tailored for traffic scenario understanding, which encompasses a rich collection of scene-level, event-level, and instance-level perception and reasoning samples. Morover, we propose RoadMind, a MLLM specifically designed for roadside interaction perception, which integrates two key components: CogniAnchor Fusion (CAF), inspired by human visual cognition for anchoring relevant regions, and Assisted Decoupled Chain-of-Thought (AD-CoT), a reasoning strategy that enhances contextual understanding and inference.
Overall, these components significantly improve VQA performance in various complex traffic environments. Our results demonstrate that RoadSceneVQA serves as a challenging and effective benchmark for advancing interactive scene understanding in real-world traffic perception tasks.



\appendix




\bibliography{aaai2026}

\begin{thebibliography}{51}
\providecommand{\natexlab}[1]{#1}

\bibitem[{Anderson et~al.(2016)Anderson, Fernando, Johnson, and Gould}]{anderson2016spice}
Anderson, P.; Fernando, B.; Johnson, M.; and Gould, S. 2016.
\newblock Spice: Semantic propositional image caption evaluation.
\newblock In \emph{European conference on computer vision}, 382--398. Springer.

\bibitem[{Anderson et~al.(2018)Anderson, He, Buehler, Teney, Johnson, Gould, and Zhang}]{anderson2018bottom}
Anderson, P.; He, X.; Buehler, C.; Teney, D.; Johnson, M.; Gould, S.; and Zhang, L. 2018.
\newblock Bottom-up and top-down attention for image captioning and visual question answering.
\newblock In \emph{Proceedings of the IEEE conference on computer vision and pattern recognition}, 6077--6086.

\bibitem[{Antol et~al.(2015)Antol, Agrawal, Lu, Mitchell, Batra, Zitnick, and Parikh}]{antol2015vqa}
Antol, S.; Agrawal, A.; Lu, J.; Mitchell, M.; Batra, D.; Zitnick, C.~L.; and Parikh, D. 2015.
\newblock Vqa: Visual question answering.
\newblock In \emph{Proceedings of the IEEE international conference on computer vision}, 2425--2433.

\bibitem[{Bai et~al.(2025)Bai, Chen, Liu, Wang, Ge, Song, Dang, Wang, Wang, Tang et~al.}]{bai2025qwen2}
Bai, S.; Chen, K.; Liu, X.; Wang, J.; Ge, W.; Song, S.; Dang, K.; Wang, P.; Wang, S.; Tang, J.; et~al. 2025.
\newblock Qwen2. 5-vl technical report.
\newblock \emph{arXiv preprint arXiv:2502.13923}.

\bibitem[{Bejarbaneh, Du, and Naghdy(2024)}]{bejarbaneh2024exploring}
Bejarbaneh, E.~Y.; Du, H.; and Naghdy, F. 2024.
\newblock Exploring shared perception and control in cooperative vehicle-intersection systems: A review.
\newblock \emph{IEEE transactions on intelligent transportation systems}.

\bibitem[{Chang et~al.(2023)Chang, Zhang, Zhang, Zhong, Peng, Li, and Li}]{chang2023bev}
Chang, C.; Zhang, J.; Zhang, K.; Zhong, W.; Peng, X.; Li, S.; and Li, L. 2023.
\newblock BEV-V2X: Cooperative birds-eye-view fusion and grid occupancy prediction via V2X-based data sharing.
\newblock \emph{IEEE Transactions on Intelligent Vehicles}, 8(11): 4498--4514.

\bibitem[{Chen et~al.(2025)Chen, Li, Zhang, Liu, Li, Gao, Hong, Tian, Zhao, Li et~al.}]{chen2025automated}
Chen, K.; Li, Y.; Zhang, W.; Liu, Y.; Li, P.; Gao, R.; Hong, L.; Tian, M.; Zhao, X.; Li, Z.; et~al. 2025.
\newblock Automated evaluation of large vision-language models on self-driving corner cases.
\newblock In \emph{2025 IEEE/CVF Winter Conference on Applications of Computer Vision (WACV)}, 7817--7826. IEEE.

\bibitem[{Chen et~al.(2024)Chen, Wu, Wang, Su, Chen, Xing, Zhong, Zhang, Zhu, Lu et~al.}]{chen2024internvl}
Chen, Z.; Wu, J.; Wang, W.; Su, W.; Chen, G.; Xing, S.; Zhong, M.; Zhang, Q.; Zhu, X.; Lu, L.; et~al. 2024.
\newblock Internvl: Scaling up vision foundation models and aligning for generic visual-linguistic tasks.
\newblock In \emph{Proceedings of the IEEE/CVF conference on computer vision and pattern recognition}, 24185--24198.

\bibitem[{Chu et~al.(2024)Chu, Qiao, Zhang, Xu, Wei, Yang, Sun, Hu, Lin, Zhang et~al.}]{chu2024mobilevlm}
Chu, X.; Qiao, L.; Zhang, X.; Xu, S.; Wei, F.; Yang, Y.; Sun, X.; Hu, Y.; Lin, X.; Zhang, B.; et~al. 2024.
\newblock Mobilevlm v2: Faster and stronger baseline for vision language model.
\newblock \emph{arXiv preprint arXiv:2402.03766}.

\bibitem[{Dai et~al.(2025)Dai, Wu, Wu, Wang, Lyu, Liao, Yu, Ding, Guan, and Yue}]{dai2025large}
Dai, W.; Wu, S.; Wu, W.; Wang, Z.; Lyu, S.; Liao, H.; Yu, L.; Ding, W.; Guan, R.; and Yue, Y. 2025.
\newblock Large Foundation Models for Trajectory Prediction in Autonomous Driving: A Comprehensive Survey.
\newblock \emph{arXiv preprint arXiv:2509.10570}.

\bibitem[{Deruyttere et~al.(2019)Deruyttere, Vandenhende, Grujicic, Van~Gool, and Moens}]{deruyttere2019talk2car}
Deruyttere, T.; Vandenhende, S.; Grujicic, D.; Van~Gool, L.; and Moens, M.~F. 2019.
\newblock Talk2Car: Taking Control of Your Self-Driving Car.
\newblock In \emph{Proceedings of the 2019 Conference on Empirical Methods in Natural Language Processing and the 9th International Joint Conference on Natural Language Processing (EMNLP-IJCNLP)}, 2088--2098.

\bibitem[{Gopalkrishnan, Greer, and Trivedi(2024)}]{gopalkrishnanmulti}
Gopalkrishnan, A.; Greer, R.; and Trivedi, M. 2024.
\newblock Multi-Frame, Lightweight \& Efficient Vision-Language Models for Question Answering in Autonomous Driving.
\newblock In \emph{First Vision and Language for Autonomous Driving and Robotics Workshop}.

\bibitem[{Guan et~al.(2025{\natexlab{a}})Guan, Hu, Zhou, Xue, Man, Smith, Lim, Ding, and Yue}]{guan2025referring}
Guan, R.; Hu, R.; Zhou, Z.; Xue, T.; Man, K.~L.; Smith, J.; Lim, E.~G.; Ding, W.; and Yue, Y. 2025{\natexlab{a}}.
\newblock Referring flexible image restoration.
\newblock \emph{Expert Systems with Applications}, 274: 126857.

\bibitem[{Guan et~al.(2025{\natexlab{b}})Guan, Jia, Yao, Yang, Xu, Purwanto, Zhu, Man, Lim, Smith et~al.}]{guan2025watervg}
Guan, R.; Jia, L.; Yao, S.; Yang, F.; Xu, S.; Purwanto, E.; Zhu, X.; Man, K.~L.; Lim, E.~G.; Smith, J.; et~al. 2025{\natexlab{b}}.
\newblock Watervg: Waterway visual grounding based on text-guided vision and mmwave radar.
\newblock \emph{IEEE Transactions on Intelligent Transportation Systems}.

\bibitem[{Guan et~al.(2024)Guan, Man, Chen, Yao, Hu, Zhu, Smith, Lim, and Yue}]{guan2024findvehicle}
Guan, R.; Man, K.~L.; Chen, F.; Yao, S.; Hu, R.; Zhu, X.; Smith, J.; Lim, E.~G.; and Yue, Y. 2024.
\newblock FindVehicle and VehicleFinder: a NER dataset for natural language-based vehicle retrieval and a keyword-based cross-modal vehicle retrieval system.
\newblock \emph{Multimedia Tools and Applications}, 83(8): 24841--24874.

\bibitem[{Guan et~al.(2023)Guan, Man, Zhao, Zhang, Yao, Smith, Lim, and Yue}]{guan2023man}
Guan, R.; Man, K.~L.; Zhao, H.; Zhang, R.; Yao, S.; Smith, J.; Lim, E.~G.; and Yue, Y. 2023.
\newblock MAN and CAT: mix attention to nn and concatenate attention to YOLO.
\newblock \emph{The Journal of Supercomputing}, 79(2): 2108--2136.

\bibitem[{Guan et~al.(2025{\natexlab{c}})Guan, Zhang, Ouyang, Liu, Man, Cai, Xu, Smith, Lim, Yue et~al.}]{guan2024talk2radar}
Guan, R.; Zhang, R.; Ouyang, N.; Liu, J.; Man, K.~L.; Cai, X.; Xu, M.; Smith, J.; Lim, E.~G.; Yue, Y.; et~al. 2025{\natexlab{c}}.
\newblock Talk2radar: Bridging natural language with 4d mmwave radar for 3d referring expression comprehension.
\newblock \emph{2025 IEEE International Conference on Robotics and Automation (ICRA)}.

\bibitem[{Han et~al.(2024)Han, Pu, Xia, Han, Pan, Li, Lu, Song, and Huang}]{han2024bridging}
Han, D.; Pu, Y.; Xia, Z.; Han, Y.; Pan, X.; Li, X.; Lu, J.; Song, S.; and Huang, G. 2024.
\newblock Bridging the divide: Reconsidering softmax and linear attention.
\newblock \emph{Advances in Neural Information Processing Systems}, 37: 79221--79245.

\bibitem[{Huang et~al.(2023)Huang, Liu, Zhou, Nguyen, Azghadi, Xia, Han, and Sun}]{huang2023v2x}
Huang, T.; Liu, J.; Zhou, X.; Nguyen, D.~C.; Azghadi, M.~R.; Xia, Y.; Han, Q.-L.; and Sun, S. 2023.
\newblock V2X cooperative perception for autonomous driving: Recent advances and challenges.
\newblock \emph{arXiv preprint arXiv:2310.03525}.

\bibitem[{Hurst et~al.(2024)Hurst, Lerer, Goucher, Perelman, Ramesh, Clark, Ostrow, Welihinda, Hayes, Radford et~al.}]{hurst2024gpt}
Hurst, A.; Lerer, A.; Goucher, A.~P.; Perelman, A.; Ramesh, A.; Clark, A.; Ostrow, A.; Welihinda, A.; Hayes, A.; Radford, A.; et~al. 2024.
\newblock Gpt-4o system card.
\newblock \emph{arXiv preprint arXiv:2410.21276}.

\bibitem[{Kim et~al.(2025)Kim, Kim, Lee, and Jang}]{kim2025visual}
Kim, B.~S.; Kim, J.; Lee, D.; and Jang, B. 2025.
\newblock Visual question answering: A survey of methods, datasets, evaluation, and challenges.
\newblock \emph{ACM Computing Surveys}, 57(10): 1--35.

\bibitem[{Krishna et~al.(2017)Krishna, Zhu, Groth, Johnson, Hata, Kravitz, Chen, Kalantidis, Li, Shamma et~al.}]{krishna2017visual}
Krishna, R.; Zhu, Y.; Groth, O.; Johnson, J.; Hata, K.; Kravitz, J.; Chen, S.; Kalantidis, Y.; Li, L.-J.; Shamma, D.~A.; et~al. 2017.
\newblock Visual genome: Connecting language and vision using crowdsourced dense image annotations.
\newblock \emph{International journal of computer vision}, 123: 32--73.

\bibitem[{Lavie and Agarwal(2007)}]{meteor}
Lavie, A.; and Agarwal, A. 2007.
\newblock Meteor.
\newblock 228--231.

\bibitem[{Liang et~al.(2025)Liang, Guan, Lian, Liu, Sun, Wu, Yue, Ding, and Xiong}]{liang2025cognitive}
Liang, S.; Guan, R.; Lian, W.; Liu, D.; Sun, X.; Wu, D.; Yue, Y.; Ding, W.; and Xiong, H. 2025.
\newblock Cognitive Disentanglement for Referring Multi-Object Tracking.
\newblock \emph{Information Fusion}, 103349.

\bibitem[{Liao et~al.(2025)Liao, Kong, Wang, Wang, Ye, He, Xu, and Li}]{liao2025cot}
Liao, H.; Kong, H.; Wang, B.; Wang, C.; Ye, W.; He, Z.; Xu, C.; and Li, Z. 2025.
\newblock Cot-drive: Efficient motion forecasting for autonomous driving with llms and chain-of-thought prompting.
\newblock \emph{IEEE Transactions on Artificial Intelligence}.

\bibitem[{Lin(2004)}]{lin2004rouge}
Lin, C.-Y. 2004.
\newblock Rouge: A package for automatic evaluation of summaries.
\newblock In \emph{Text summarization branches out}, 74--81.

\bibitem[{Malinowski, Rohrbach, and Fritz(2015)}]{malinowski2015ask}
Malinowski, M.; Rohrbach, M.; and Fritz, M. 2015.
\newblock Ask your neurons: A neural-based approach to answering questions about images.
\newblock In \emph{Proceedings of the IEEE international conference on computer vision}, 1--9.

\bibitem[{Muhammad et~al.(2022)Muhammad, Hussain, Ullah, Del~Ser, Rezaei, Kumar, Hijji, Bellavista, and de~Albuquerque}]{muhammad2022vision}
Muhammad, K.; Hussain, T.; Ullah, H.; Del~Ser, J.; Rezaei, M.; Kumar, N.; Hijji, M.; Bellavista, P.; and de~Albuquerque, V. H.~C. 2022.
\newblock Vision-based semantic segmentation in scene understanding for autonomous driving: Recent achievements, challenges, and outlooks.
\newblock \emph{IEEE Transactions on Intelligent Transportation Systems}, 23(12): 22694--22715.

\bibitem[{Nam, Ha, and Kim(2017)}]{nam2017dual}
Nam, H.; Ha, J.-W.; and Kim, J. 2017.
\newblock Dual attention networks for multimodal reasoning and matching.
\newblock In \emph{Proceedings of the IEEE conference on computer vision and pattern recognition}, 299--307.

\bibitem[{Papineni et~al.(2002)Papineni, Roukos, Ward, and Zhu}]{papineni2002bleu}
Papineni, K.; Roukos, S.; Ward, T.; and Zhu, W.-J. 2002.
\newblock Bleu: a method for automatic evaluation of machine translation.
\newblock In \emph{Proceedings of the 40th annual meeting of the Association for Computational Linguistics}, 311--318.

\bibitem[{Qian et~al.(2024)Qian, Chen, Zhuo, Jiao, and Jiang}]{qian2024nuscenes}
Qian, T.; Chen, J.; Zhuo, L.; Jiao, Y.; and Jiang, Y.-G. 2024.
\newblock Nuscenes-qa: A multi-modal visual question answering benchmark for autonomous driving scenario.
\newblock In \emph{Proceedings of the AAAI Conference on Artificial Intelligence}, volume~38, 4542--4550.

\bibitem[{Sima et~al.(2024)Sima, Renz, Chitta, Chen, Zhang, Xie, Bei{\ss}wenger, Luo, Geiger, and Li}]{sima2024drivelm}
Sima, C.; Renz, K.; Chitta, K.; Chen, L.; Zhang, H.; Xie, C.; Bei{\ss}wenger, J.; Luo, P.; Geiger, A.; and Li, H. 2024.
\newblock Drivelm: Driving with graph visual question answering.
\newblock In \emph{European Conference on Computer Vision}, 256--274. Springer.

\bibitem[{Sun et~al.(2024)Sun, Song, Liu, Yang, Wang, Li, Yang, and Chu}]{sun20243d}
Sun, P.; Song, Y.; Liu, X.; Yang, X.; Wang, Q.; Li, T.; Yang, Y.; and Chu, X. 2024.
\newblock 3d question answering for city scene understanding.
\newblock In \emph{Proceedings of the 32nd ACM International Conference on Multimedia}, 2156--2165.

\bibitem[{Sun et~al.(2025)Sun, Song, Zhu, Liu, Wang, Liu, Xia, Li, Yang, and Chu}]{sun2025city}
Sun, P.; Song, Y.; Zhu, X.; Liu, X.; Wang, Q.; Liu, Y.; Xia, C.; Li, T.; Yang, Y.; and Chu, X. 2025.
\newblock City-VLM: Towards Multidomain Perception Scene Understanding via Multimodal Incomplete Learning.
\newblock In \emph{Proceedings of the 33rd ACM International Conference on Multimedia}, 3448--3457.

\bibitem[{Vaswani et~al.(2017)Vaswani, Shazeer, Parmar, Uszkoreit, Jones, Gomez, Kaiser, and Polosukhin}]{vaswani2017attention}
Vaswani, A.; Shazeer, N.; Parmar, N.; Uszkoreit, J.; Jones, L.; Gomez, A.~N.; Kaiser, {\L}.; and Polosukhin, I. 2017.
\newblock Attention is all you need.
\newblock \emph{Advances in neural information processing systems}, 30.

\bibitem[{Vedantam, Lawrence~Zitnick, and Parikh(2015)}]{vedantam2015cider}
Vedantam, R.; Lawrence~Zitnick, C.; and Parikh, D. 2015.
\newblock Cider: Consensus-based image description evaluation.
\newblock In \emph{Proceedings of the IEEE conference on computer vision and pattern recognition}, 4566--4575.

\bibitem[{Wang et~al.(2020)Wang, Li, Khabsa, Fang, and Ma}]{wang2020linformer}
Wang, S.; Li, B.~Z.; Khabsa, M.; Fang, H.; and Ma, H. 2020.
\newblock Linformer: Self-attention with linear complexity.
\newblock \emph{arXiv preprint arXiv:2006.04768}.

\bibitem[{Wu et~al.(2025)Wu, Han, Liu, Wang, Xu, Zhang, and Shen}]{wu2025language}
Wu, D.; Han, W.; Liu, Y.; Wang, T.; Xu, C.-z.; Zhang, X.; and Shen, J. 2025.
\newblock Language prompt for autonomous driving.
\newblock In \emph{Proceedings of the AAAI Conference on Artificial Intelligence}, volume~39, 8359--8367.

\bibitem[{Xie et~al.(2025)Xie, Kong, Dong, Sima, Zhang, Chen, Liu, and Pan}]{xie2025vlms}
Xie, S.; Kong, L.; Dong, Y.; Sima, C.; Zhang, W.; Chen, Q.~A.; Liu, Z.; and Pan, L. 2025.
\newblock Are VLMs Ready for Autonomous Driving? An Empirical Study from the Reliability, Data, and Metric Perspectives.
\newblock \emph{arXiv preprint arXiv:2501.04003}.

\bibitem[{Xu et~al.(2021)Xu, Berres, Tennille, Ravulaparthy, Wang, and Sanyal}]{xu2021continuous}
Xu, H.; Berres, A.; Tennille, S.~A.; Ravulaparthy, S.~K.; Wang, C.; and Sanyal, J. 2021.
\newblock Continuous emulation and multiscale visualization of traffic flow using stationary roadside sensor data.
\newblock \emph{IEEE Transactions on Intelligent Transportation Systems}, 23(8): 10530--10541.

\bibitem[{Yang et~al.(2024)Yang, Zhang, Wang, and Xie}]{yang2024mma}
Yang, L.; Zhang, R.-Y.; Wang, Y.; and Xie, X. 2024.
\newblock Mma: Multi-modal adapter for vision-language models.
\newblock In \emph{Proceedings of the IEEE/CVF Conference on Computer Vision and Pattern Recognition}, 23826--23837.

\bibitem[{Yang et~al.(2016)Yang, He, Gao, Deng, and Smola}]{yang2016stacked}
Yang, Z.; He, X.; Gao, J.; Deng, L.; and Smola, A. 2016.
\newblock Stacked attention networks for image question answering.
\newblock In \emph{Proceedings of the IEEE conference on computer vision and pattern recognition}, 21--29.

\bibitem[{Yao et~al.(2024)Yao, Yu, Zhang, Wang, Cui, Zhu, Cai, Li, Zhao, He et~al.}]{yao2024minicpm}
Yao, Y.; Yu, T.; Zhang, A.; Wang, C.; Cui, J.; Zhu, H.; Cai, T.; Li, H.; Zhao, W.; He, Z.; et~al. 2024.
\newblock Minicpm-v: A gpt-4v level mllm on your phone.
\newblock \emph{arXiv preprint arXiv:2408.01800}.

\bibitem[{Ye et~al.(2022)Ye, Shu, Li, Shi, Li, Wang, Tan, and Ding}]{ye2022rope3d}
Ye, X.; Shu, M.; Li, H.; Shi, Y.; Li, Y.; Wang, G.; Tan, X.; and Ding, E. 2022.
\newblock Rope3d: The roadside perception dataset for autonomous driving and monocular 3d object detection task.
\newblock In \emph{Proceedings of the IEEE/CVF Conference on Computer Vision and Pattern Recognition}, 21341--21350.

\bibitem[{Yu et~al.(2022)Yu, Luo, Shu, Huo, Yang, Shi, Guo, Li, Hu, Yuan et~al.}]{yu2022dair}
Yu, H.; Luo, Y.; Shu, M.; Huo, Y.; Yang, Z.; Shi, Y.; Guo, Z.; Li, H.; Hu, X.; Yuan, J.; et~al. 2022.
\newblock Dair-v2x: A large-scale dataset for vehicle-infrastructure cooperative 3d object detection.
\newblock In \emph{Proceedings of the IEEE/CVF Conference on Computer Vision and Pattern Recognition}, 21361--21370.

\bibitem[{Zhang et~al.(2025{\natexlab{a}})Zhang, Zhou, Zheng, Gao, Cui, Li, Chen, and Zhang}]{zhang2025open3dvqa}
Zhang, W.; Zhou, Z.; Zheng, Z.; Gao, C.; Cui, J.; Li, Y.; Chen, X.; and Zhang, X.-P. 2025{\natexlab{a}}.
\newblock Open3dvqa: A benchmark for comprehensive spatial reasoning with multimodal large language model in open space.
\newblock \emph{arXiv preprint arXiv:2503.11094}.

\bibitem[{Zhang et~al.(2025{\natexlab{b}})Zhang, Li, Xu, Peng, Zhou, Shi, and Huang}]{zhang2025mpdrive}
Zhang, Z.; Li, X.; Xu, Z.; Peng, W.; Zhou, Z.; Shi, M.; and Huang, S. 2025{\natexlab{b}}.
\newblock Mpdrive: Improving spatial understanding with marker-based prompt learning for autonomous driving.
\newblock In \emph{Proceedings of the Computer Vision and Pattern Recognition Conference}, 12089--12099.

\bibitem[{Zhang et~al.(2024)Zhang, Zhang, Li, Zhao, Karypis, and Smola}]{zhang2024multimodal}
Zhang, Z.; Zhang, A.; Li, M.; Zhao, H.; Karypis, G.; and Smola, A. 2024.
\newblock Multimodal Chain-of-Thought Reasoning in Language Models.
\newblock \emph{Transactions on Machine Learning Research}, 2024.

\bibitem[{Zhou et~al.(2024{\natexlab{a}})Zhou, Hu, Weng, Jia, Luo, Liu, Wu, and Huang}]{zhou2024tinyllava}
Zhou, B.; Hu, Y.; Weng, X.; Jia, J.; Luo, J.; Liu, X.; Wu, J.; and Huang, L. 2024{\natexlab{a}}.
\newblock Tinyllava: A framework of small-scale large multimodal models.
\newblock \emph{arXiv preprint arXiv:2402.14289}.

\bibitem[{Zhou et~al.(2025)Zhou, Larintzakis, Guo, Zimmer, Liu, Cao, Zhang, Lakshminarasimhan, Strand, and Knoll}]{zhou2025tumtraffic}
Zhou, X.; Larintzakis, K.; Guo, H.; Zimmer, W.; Liu, M.; Cao, H.; Zhang, J.; Lakshminarasimhan, V.; Strand, L.; and Knoll, A.~C. 2025.
\newblock TUMTraffic-VideoQA: A Benchmark for Unified Spatio-Temporal Video Understanding in Traffic Scenes.
\newblock \emph{arXiv preprint arXiv:2502.02449}.

\bibitem[{Zhou et~al.(2024{\natexlab{b}})Zhou, Liu, Yurtsever, Zagar, Zimmer, Cao, and Knoll}]{zhou2024vision}
Zhou, X.; Liu, M.; Yurtsever, E.; Zagar, B.~L.; Zimmer, W.; Cao, H.; and Knoll, A.~C. 2024{\natexlab{b}}.
\newblock Vision language models in autonomous driving: A survey and outlook.
\newblock \emph{IEEE Transactions on Intelligent Vehicles}.

\end{thebibliography}

\end{document}